\pdfoutput=1

\documentclass[11pt]{article}

\usepackage{acl}
\usepackage{times}
\usepackage{latexsym}
\usepackage[mathscr]{euscript}
\usepackage[T1]{fontenc}
\usepackage{multirow}
\usepackage[utf8]{inputenc}
\usepackage{amsmath} 
\usepackage{microtype}
\usepackage{amssymb}
\usepackage{color}
\usepackage{graphicx}
\usepackage{arydshln}
\usepackage{float}
\usepackage{stfloats}
\usepackage[title]{appendix}
\usepackage{setspace}
\title{StyleFlow: Disentangle Latent Representations via Normalizing Flow for Unsupervised Text Style Transfer}

\author{Kangchen Zhu, Zhiliang Tian, Ruifeng Luo, Xiaoguang Mao \\
    National University of Defense Technology \\
  \texttt{\{zhukangchen18, tianzhiliang, luoruifeng21, xgmao\}@nudt.edu.cn} 
  }

\begin{document}
\maketitle
\begin{abstract}
Text style transfer aims to alter the style of a sentence while preserving its content. Due to the lack of parallel corpora, most recent work focuses on unsupervised methods and often uses cycle construction to train models. Since cycle construction helps to improve the style transfer ability of the model by rebuilding transferred sentences back to original-style sentences, it brings about a content loss in unsupervised text style transfer tasks. In this paper, we propose a novel disentanglement-based style transfer model StyleFlow to enhance content preservation. Instead of the typical encoder-decoder scheme, StyleFlow can not only conduct the forward process to obtain the output, but also infer to the input through the output. We design an attention-aware coupling layers to disentangle the content representations and the style representations of a sentence. Besides, we propose a data augmentation method based on Normalizing Flow to improve the robustness of the model. Experiment results demonstrate that our model preserves content effectively and achieves the state-of-the-art performance on the most metrics.

\end{abstract}

\section{Introduction}

Style transfer is a popular task in natural language processing, referring to the process of converting a given sentence with a certain style (e.g., sentiment) into another style while retaining the original content \cite{shen2017style,fu2018style,john2018disentangled,lee2021enhancing}. Recently, deep learning have become the dominant methods in text style transfer. Due to the lack of large parallel corpora, the majority of recent researches have focused on unsupervised learning and generally reach to this purpose by combining  representations of source content and target styles\cite{fu2018style}.

One mainstream approach is to encode sentences to obtain full content information, and generate target sentences by manipulating style attributes during decoding. \citet{lample2018multiple} take this approach and try to control the attributes of text with an attention-based LSTM model. Instead of LSTM, \citet{dai2019style} utilize the Transformer\cite{vaswani2017attention} model to encode the source sentences and change the style embedding to control the style of target sentences. This method can maintain the content information well but does not effectively remove the style attributes from the original sentences, which leads to that the generated sentences are not stylistic.

Another prevalent paradigm is a disentanglement approach: it explicitly disentangles the sentence into content representations and style representations, and then combines source content and target style to generate a target-style sentence. \citet{shen2017style} disentangle the encoded vectors into content and style vectors by using a cross-aligned auto-encoder. \citet{john2018disentangled} use denoising auto-encoding \cite{vincent2010stacked} and variational auto-encoder \cite{kingma2013auto} to disentangle the content and style space to avoid auto-encoder copying from the input directly. \citet{fu2018style} propose to disentangle separate content and style representations using adversarial networks \cite{goodfellow2014generative}. However, some researchers have observed that words or semantic tokens carry both style and content information \cite{li2018delete,yi2021text}. For instance, the word "delicious" conveys strong style information while also implying a food-related topic. Such words serve as an indicator of style (e.g., positive sentiment) and content(e.g., food) information at the same time. Thus, the above methods are incapable to disentangle completely the content and style space, which leads to the loss of content information. 

As the models cannot achieve perfect disentanglement, the models may miss some content information while conducting "cycle reconstruction". \cite{dos2018fighting,logeswaran2018content,luo2019dual,dai2019style,yi2021text}. Cycle reconstruction is a very common strategy to maintain the content information in unsupervised style transfer, where parallel supervised samples are not accessible. It reconstructs the transferred sentences back to original-style sentences. Specifically, most cycle reconstruction methods \cite{luo2019dual,dai2019style,yi2021text} cannot completely reconstruct the original inputs (i.e. cycle loss is not zero during the training), causing the model to generate target-style sentences with content information missing.

In this paper, we propose a novel disentanglement based style transfer model \textit{StyleFlow} that maintains content information well. Specifically, StyleFlow takes Normalizing Flow \cite{rezende2015variational} as the basic block. Normalizing Flow is a flow-based symmetric reversible neural network component \cite{dinh2014nice,kingma2018glow,rezende2015variational}. Different from traditional encoder-decoder style transfer models, Normalizing Flow encodes in the forward process and decodes in the reverse process without additional loss \cite{yuksel2021semantic}.
In cycle reconstruction, StyleFlow applies coupling layers to completely rebuild the transferred sentences back with original style without any content loss (e.g. cycle loss = 0) owing to its reversibility. We equip coupling layers with attention mechanism
 \cite{vaswani2017attention} at token level \cite{yi2021text} to disentangle the words carrying both strong styles and contents information for better content preservation. 
In addition, to enhance the robustness of the model, we propose a data augmentation method based on Normalizing Flow. It adds perturbations to the encoded vectors, and then decodes pseudo sentences with the same style, which locate in the neighborhood of the original sentence. Experiment results show that our model is capable to  preserve content effectively and achieves the state-of-the-art performance on the most metrics.

Our contributions are as follows:
\begin{itemize}
\item  We propose a novel disentanglement method StyleFlow, which takes Normalizing Flow as a basic block. Its reversibility guarantees no content loss during the forward process (encoding) and reverse process (decoding). 
\item  We apply the attention-aware coupling layer to fully reconstruct the transferred sentences back with original style without any content loss. Besides, we propose a data augmentation method based on Normalizing Flow to improve the robustness of the model.
\item  Experiments indicate that our proposed model outperforms other baselines on two datasets. Specifically, StyleFlow achieves state-of-the-art in most metrics.
\end{itemize}

\section{Related Work}
\subsection{Text Style Transfer} 
In recent years, researchers have been studying style transfer in natural language processing. In text style transfer, stylistic properties are changed (such as sentiment) while maintaining source content. 

A mainstream approach is to encode sentences in order to obtain complete content information, and then decode the sentences using style attributes to generate target sentences. \citet{lample2018multiple} take this approach and try to control the attributes of text with an attention-based LSTM model. Instead of LSTM, \citet{dai2019style} utilize the Transformer\cite{vaswani2017attention} model to encode the source sentences and change the style embedding to control the style of target sentences. Since the learned embedded model can not adequately represent the highly complex concept of style, this paradigm usually results in unsatisfactory results when it comes to style transfer. 

Another prevalent methods attempt to disentangle the sentence into  style space and content space \cite{shen2017style, li2018delete, fu2018style}. \citet{shen2017style} propose a cross-aligned auto-encoder with adversarial training to learn a shared latent content distribution and a separated latent style distribution. \citet{fu2018style} concatenate the extracted content with a learned target-style embedding. \citet{john2018disentangled} design multiple losses to pack a sentence into a latent space, which is then split into sub-spaces of content and style. Since complete disentanglement is impracticable, this paradigm usually results in satisfactory style accuracy but poor content preservation.

\subsection{Normalizing Flow} 
Normalizing Flow (NF) is a family of generative models with tractable distributions where both sampling and density evaluation can be efficient and exact. Specifically, NF is a transformation of a simple probability distribution (e.g., a standard Gaussian distribution) into a more complex distribution by a sequence of invertible and differentiable mappings.

In Computer Vision area, Normalizing Flow has been widely used for image generation tasks. \citet{kingma2018glow} propose Glow generative Flow with invertible $1\times1$ convolutions for image generation. Based on Glow, \citet{nagar2021cinc} propose characterizable invertible $3\times3$ Convolution Flow method to generate images effectively. \citet{an2021artflow} propose ArtFlow to achieve unbiased image style transfer via reversible neural flows. Nowadays researchers have paid more attention to use Normalizing Flow technique to solve the tasks in Natural Language Processing. \citet{li2020sentence} use normalizing flow to calculate the cosine similarity of embeddings from BERT \cite{devlin2018bert}. \citet{ma2019flowseq} propose Flowseq to achieve non-autoregressive conditional sequence generation with generative flow on neural machine translation.

In this paper, we propose a novel disentanglement based style transfer model \textit{StyleFlow}, which applies Normalizing Flow in text style transfer tasks to enhance content preservation.

\section{Task Definition}
Let $D=(\mathbf{x}_{i}, s_{i})$ represents a training corpus, with $\mathbf{x}_{i}$ representing sentences and $\mathbf{x}_{i}$ representing style labels. We conducted our experiments on a sentiment analysis task, which has two style labels ("positive" and "negative") .
The task is to learn from $D$ a model $\hat{\mathbf{x}}_{\hat{s}} = f_{\theta}(\mathbf{x},\hat{s})$ with parameters $\theta$, which inputs a source sentence $\mathbf{x}$ and a target style $\hat{s}$ and outputs a new sentence $\hat{\mathbf{x}}_{\hat{s}}$ that has the target style and maintains the content information of $\mathbf{x}$.

\section{StyleFlow Network}
In this work, we propose an innovative method \textit{StyleFlow} to address the problem of content loss in style transfer task. Different from the commonly used encoder-decoder schema, we use Normalizing Flow to build a flow-based reversible encoder. The special encoder can rebuild the input from the output, which allows it to encode the source sentence in the forward process and generate sentences in the reverse process. With its lossless encoder, StyleFlow can reduce the loss of content information in style transfers. 

In this section, we first introduce the principle of the flow-based network in Section 4.1, discuss the use of the coupling layer structure in text style transfer in Section 4.2, discuss the details of the flow-based reversible encoder in Section 4.3, and discuss the objective functions of model training in Section 4.4.

\subsection{Flow-based Generative Network}
Flow-based generative network refers to a subclass of deep generative models, which learns the exact likelihood of high dimensional observations(e.g., natural images, texts, and audios) through a chain of reversible transformations. \citet{rezende2015variational} proposed the idea of Normalizing Flows, which is one of flow-based generative network with a sequence of invertible transformations applied to probability distributions to obtain a probability distribution with the desired properties.

Consider a smooth invertible mapping $f:\mathbb{R}^{d}\to\mathbb{R}^{d}$ where $d$ is the dimensionality of the latent space. Then, we can transform the random variable $\mathbf{z}\sim q(\mathbf{z})$ by using the mapping $f$. This transformation returns the random variable $\mathbf{z^{\prime}}=f(\mathbf{z})$ with the following distribution.
\begin{equation}
q(z^{\prime})=q(z)\left |det\frac{\partial f^{-1} }{\partial z^{\prime}} \right |=q(z)\left |det\frac{\partial f}{\partial z^{\prime}} \right |^{-1}
\label{formula 1}
\end{equation}

Because the function $f$ is invertible, this result follows from the chain rule and is applicable to the Jacobian of invertible functions using the inverse function theorem.

We can extend this idea to a sequence of $K$ transformations $f_{k}$ and transform a random variable $\mathbf{z}_{0}\sim q_{0}$ to obtain a transformed density $q_K(\mathbf{z}):$
\begin{equation}
\mathbf{z}_{K}=f_{K}\circ \cdots\circ f_{2}\circ f_{1}
\label{formula 2}
\end{equation}
\begin{small}
\begin{equation}
log q_{K}(\mathbf{z}_{K})=log q_{0}(\mathbf{z}_{0})- {\textstyle \sum_{k=1}^{K}} log\left | det\frac{\partial f_{k}}{\partial \mathbf{z}_{k-1}}  \right | 
\label{formula 3}
\end{equation}
\end{small}
Where $\circ$ denotes function composition.

\subsection{Coupling Layer}
\citet{dinh2014nice} proposed an expressive reversible transformation named coupling layer. A special case of coupling layer is additive coupling layer and the forward computation of additive coupling layer is
\begin{equation}
\begin{split}
    &x_a,x_b=split(x)\\
    &s, t=NN(x_b)\\
    &y_a = s\odot x_a + t \\
    &y_b = x_b \\
    &y=concat(y_a,y_b).
\end{split}
\label{formula 4}
\end{equation}
The $split()$ function splits a tensor into two halves along the channel dimension. $NN()$ is any neural network where the input and the output have the same shape. The $\odot$ is dot product operation. The $concat()$ function concatenates two tensors along the channel dimension. The reverse computation of additive coupling can be easily derived.

In text style transfer task, we pass the sentence $x$ as input into the coupling layer, which divides the sentence embeddings into two parts $x_a$ and $x_b$.
Since the neural network function $NN()$ in the coupling layer can be selected arbitrarily, we choose the Transformer \cite{dai2019style} structure that performs well in text generation tasks as the basic module. Since Transformer consists of  multi layers, we choose a single-layer Transformer block as the basic module to obtain the parameters in formula (4) :
\begin{equation}
    s, t=Transformer\_Block(x_a)\\
\label{formula 5}
\end{equation}
As mentioned in Formula \ref{formula 4} and \ref{formula 5}, assume that $x_a$ is passed into the Transformer block module to obtain $s,t$, and then concatenate $y_a$ and $y_b$ to obtain $y$.
Since only part of the sentence information is encoded (e.g. $x_a$) in the forward process of each coupling layer while another part remains the same, which leads inadequate encoding of sentences. Some researchers connect multiple coupling layers to obtain a flow-based model for better encoding effect. Since each coupling layer is reversible, the flow-based model remains reversible.

\subsection{Flow-based Reversible Encoder}
Different from the encoder-decoder scheme used in text style transfer, StyleFlow has only an encoder and works in an \textit{encoding-disentanglement-reverse} scheme. In the encoding step, the reversible encoder encodes the input sentence $x$ into a vector representation $z_x$ in the forward process. In the disentanglement step, the model disentangles the encoded vector $z$ into a content vector $z_c$ and a style vector $z_s$. In the revere step, we first generate the target style $z_{s^{\prime}}$ and then concatenate it with the source content vector $z_c$  to generate the vector representation $z^{\prime}$. Finally, the reversible encoder generates the target-style sentence $y$ in the reverse process. 

\subsubsection{Encoding}
In the encoding step, we propose the \textbf{attention-aware coupling layer}, which uses the attention mechanism on the coupling layer to improve its ability of disentanglement. Each coupling layer structure uses \textit{attention-split} method, which applies the pre-trained bidirectional GRU \cite{cho2014learning} model with attention mechanism to score the tokens of sentence $x$. 
As is shown in Formula \ref{formula 6}, we choose the tokens with scores above the threshold as the style $x_s$ and the rest tokens as content $x_c$. Then, the content tokens $x_c$ are fed into the Transformer Block to obtain the parameters $s,t$ for calculating $z_s=s\odot x_s + t$, while $x_c$ remains unchanged ($z_c=s_c$). Finally, model concatenates $z_s$ and $z_c$ to get a new representation $z$. 
\begin{equation}
\begin{split}
    &x_c,x_s=attention\_split(x)\\
    &s, t=Transformer\_Block(x_c)\\
    &z_s = s\odot x_s + t \\
    &z_c = x_c \\
    &z=concat(z_s,z_c).
\end{split}
\label{formula 6}
\end{equation}
Similarly, we feed the output of the last coupling layer $z$ to the next one to obtain the encoding vector. The model repeats the above operations in other attention-aware coupling layer, and obtains the target-style sentence $y$ through the flow-based reversible encoder.

\subsubsection{Disentanglement}
In the disentanglement step, the model aims to split $z$ into two vectors, which indicates style information and content information respectively. 
In the disentanglement step, the model aims to split $z$ into two vectors, which indicates style information and content information respectively.
Particularly, we apply the attention mechanism to score the encoded vector $z$. The score higher than the threshold is used as the style vector $z_s$, and the rest is used as the content vector $z_c$. Since in the encode step, the sentence is divided into content and style for encoding, it is easier to divide the sentence into two spaces of content and style by using the attention mechanism in the disentanglement step.

\subsubsection{Reverse}
In the reverse phrase, we use three steps to generate target style sentences. In the first step, we follow \citet{lee2021enhancing} to use conditional layer normalization approach to generate content-related representation $z_{s^{\prime}}$ of the target style:
\begin{equation}
\begin{split}
    &\quad N(z_c)= \frac{{z_c}-\mu_c}{\sigma_c} \\
    &z_{s^{\prime}}=\gamma^{s^{\prime}}\odot N(z_c)+\beta^{s^{\prime}}
\end{split}
\label{formula 7}
\end{equation}
where $\mu_c$ and $\sigma_c$ are mean and standard deviation
of content vector respectively. $s^{\prime}$ is source style label. 
The model learns separate $\gamma^{s^{\prime}}$ and $\beta^{s^{\prime}}$
parameters for different styles.

In the second step, we fuse the source content vector $z_c$ and the target style vector $z_{s^{\prime}}$ to obtain target-style representation $z^{\prime}$. 
Similar to the disentanglement method, we also use the attention mechanism to fuse two vectors $z_c$ and $z_{s^{\prime}}$. In the disentanglement step, the vectors $z$ is disentangled by the attention score into the content space $z_c$ and the style space $z_s$, so the target style vector $z_{s^{\prime}}$ should be spliced at the position of the source style vector $z_c$  to obtain the target vector $z^{\prime}$. Compared with directly concatenating two vectors, the above fusion method based on the attention mechanism does not change the previous arrangement relationship of the vectors, which is more efficient to reducing content loss when generating target-style sentences. 

In the third step, we use the reverse process of reversible encoder to generate target-style sentences. Specifically, we use the attention-split method used in Formula \ref{formula 6} to split $z^{\prime}$ into $z^{\prime}_c$ and $z^{\prime}_s$, and then pass it into the Transformer block to calculate the parameters $s, t$:
\begin{equation}
\begin{split}
    &z^{\prime}_c,z^{\prime}_s=attention\_split(z^{\prime})\\
    &s, t=Transformer\_Block(z^{\prime}_c)\\
    &{x_s}^{\prime} = (z^{\prime}_s-t)/s \\
    &{x_c}^{\prime} = z^{\prime}_c \\
    &x^{\prime}=concat({x_s}^{\prime},{x_c}^{\prime}).
\end{split}
\label{formula 8}
\end{equation}
The reverse process of coupling layer rebuilds the target-style sentence $x^{\prime}$.

\subsection{Objective Functions}
The objective function of our algorithm consists of four parts: a self reconstruction loss $\mathcal{L}_{self}$, a cycle reconstruction loss $\mathcal{L}_{cycle}$, a content loss $\mathcal{L}_{c}$, and a style transfer loss $\mathcal{L}_{s}$.
\subsubsection{Self Reconstruction Loss}
For a training example $(\mathbf{x}, s)\in \mathcal{D}$, when the sample $\mathbf{x}$ and style $s$ come from the same dataset, we would expect it to reconstruct the input sentence.
\begin{equation}
\mathcal{L}_{self}=-\mathbb{E}_{(\mathbf{x},s)\sim\mathcal{D}}[logP_{D}(\mathbf{x}|\mathbf{z_{\mathbf{x}}},\mathbf{z}_{s})]
\label{formula 9}
\end{equation}
where $\mathbf{z}_{\mathbf{x}}$ is the content representation of the input $\mathbf{x}$, and $\mathbf{z}_{s}$ is the representation of the style $s$, and $P_{D}$ is the conditional distribution. 

\subsubsection{Cycle Reconstruction Loss}
When the sample $\mathbf{x}$ and style $\hat{s}$ come from the different dataset, we transfer the sentence $\mathbf{x}$ with $\hat{s}$ to get $\hat{\mathbf{x}}_{\hat{s}}$, and then transfer $\hat{\mathbf{x}}_{\hat{s}}$ back to the original style $s$. We would expect the transferred sentences are as close as possible with the original sentences.
\begin{equation}
\mathcal{L}_{cycle}=-\mathbb{E}_{(\mathbf{x},s)\sim\mathcal{D}}[logP_{D}(\mathbf{x}|\mathbf{z}_{\hat{\mathbf{x}}_{\hat{s}}},\mathbf{z}_{s})]
\label{formula 10}
\end{equation}
where $\mathbf{z}_{\hat{\mathbf{x}}_{\hat{s}}}$ is the content representation of the target-style sentence $\hat{\mathbf{x}}_{\hat{s}}$, and $\mathbf{z}_{s}$ is the representation of the original style $s$, and $P_{D}$ is the conditional distribution. 

\subsubsection{Content Loss}
In the cycle reconstruction process, we obtain content representations $\mathbf{z}_{c}$ of the input $\mathbf{x}$ and also obtain the content representation $\mathbf{z}^{\prime}_c$ of the transferred sentence $\hat{\mathbf{x}}_{\hat{s}}$. 
We expect that only the style vector is transferred and the content vector is unchanged in style transfer, so the two representations of content should be the same. The content loss is as follows:

\begin{equation}
\mathcal{L}_{c}=\mathbb{E}_{(\mathbf{x},s)\sim\mathcal{D}}\left \|\mathbf{z}_{c}- \mathbf{z}^{\prime}_c \right \|_{2}^{2} 
\label{formula 11}
\end{equation}

\subsubsection{Style Transfer Loss}
The transferred sentences $\hat{\mathbf{x}}_{\hat{s}}$ are expected to be of the target style $\hat{s}$. The style transfer loss is as follows:
\begin{equation}
\mathcal{L}_{style}=-\mathbb{E}_{(\mathbf{x},s)\sim\mathcal{D}}[logp_{C}(\hat{s}|\hat{\mathbf{x}}_{\hat{s}})]
\label{formula 12}
\end{equation}
where $p_{C}$ is the conditional distribution over styles by the style classifier.
\subsubsection{Total Loss}
In summary, we take the four loss functions into consideration to train our model. We expect to minimum the objective function as follows:
\begin{equation}
 \mathcal{L}_{total} = \lambda_{1}\mathcal{L}_{self} +\lambda_{2}\mathcal{L}_{cycle} +\lambda_{3}\mathcal{L}_{c} +\lambda_{4}\mathcal{L}_{s}
 \label{formula 13}
\end{equation}
where $\lambda_{i}, i=1,2,3,4$ are weight parameters.
\section{Text Data Augmentation via Normalizing Flow}
In text style transfer, sample data for some styles are limited, resulting in poor robustness of the model. We propose a text data augmentation module with the Normalizing Flow structure. 

We propose to use the Normalizing Flow to establish a mapping between the sample and the latent layer space, and adds Gaussian noise to the latent space. Then, the model regenerates new samples in the neighborhood of the original samples through the reverse process. Compared with other text data enhancement methods, the samples generated by our method is more realistic to enhance the robustness of the model.

The invertibility of Normalizing Flows enables bidirectional transitions between text and latent spaces, which allows for applying perturbations
directly in the latent space rather than sample space. We recall that we denote by $\mathcal{F}:\mathcal{X}\to \mathcal{Z}$ a trained Normalizing Flow, mapping from data manifold $\mathcal{X}$ to latent space $\mathcal{Z}$. Given a perturbation function $\mathcal{F}:\mathcal{X}\to \mathcal{Z}$, defined over the latent space, we define its counterpart in sample space as $\mathcal{F}^{-1}(\mathcal{P}(\mathcal{F}(x)))$. Specifically, given a data point $x_i$, its latent representation is $z_i=\mathcal{F}(x_i)$. Then, we use the standard Gaussian distribution as the perturbation function:
\begin{equation}
P_{rand}(\cdot,\epsilon)=\Pi \left(\epsilon\cdot \mathcal{N} \left(0,\mathbf{I}\right)\right)
\end{equation}

where we use an extra parameter $\epsilon$
to control the size of perturbation. $\mathcal{P}$ is independent of $z_i$.

Any such distribution around the original $z_i$ is equivalent to sampling from the learned manifold. In this case, the Normalizing Flow pushes forward this simple Gaussian distribution centered around $z_i$
to a distribution on the sample space around $x_i = F^{-1}(z_i)$. Thus, sampling from the simple prior distribution $\mathcal{N}(0,\mathbf{I})$ is equivalent to sampling from a complex conditional distribution around the original text over the data manifold.
\section{Experiment}

\subsection{Datasets}
We evaluated and compared our approach with several state-of-the-art systems on two review datasets, Yelp Review Dataset (Yelp) and IMDb Movie Review Dataset (IMDb). The statistics of
the two datasets are shown in Table 1.

\noindent\textbf{Yelp Review Dataset (Yelp)}
The Yelp dataset is provided by the Yelp Dataset Challenge, consisting of restaurants and business reviews with sentiment labels (negative or positive). Following previous work, we use the possessed dataset provided by \cite{li2018delete}. Additionally, it also provides human reference sentences for the test set.

\noindent\textbf{IMDb Movie Review Dataset (IMDb)}
The IMDb dataset consists of movie reviews written by online users \cite{dai2019style}. This dataset is comprised
of 17.9K positive and 18.8K negative reviews for training corpus, and 2K sentences are used for testing.

\begin{table}[ht] \small 
\renewcommand\arraystretch{1.3}
\centering
\setlength{\tabcolsep}{1.8mm}{
\begin{tabular}{cccccc} 
\hline
\multirow{2}{*}{Dataset} & \multicolumn{2}{c}{Yelp} &  & \multicolumn{2}{c}{IMDb} \\ \cline{2-3} \cline{5-6}
                         & Positive    & Negative    &  & Positive    & Negative    \\ \hline
Train                    & 266,041     & 177,218    &  & 178,869     & 187,597    \\
Dev                      & 2,000       & 2,000      &  & 2,000       & 2,000      \\
Test                     & 5,00        & 5,00       &  & 1,000       & 1,000      \\ \cline{2-3} \cline{5-6}
Avg.Len.                 & \multicolumn{2}{c}{8.9}  &  & \multicolumn{2}{c}{18.5} \\ \hline

\end{tabular}}
\caption{\label{Table1}
The statistics of the Yelp and IMDb datasets.
}
\end{table}
\subsection{Baselines}
We conduct comprehensive comparisons with several state-of-the-art style transfer models. For unsupervised transfer, we consider CrossAlign \cite{shen2017style}, ControlledGen \cite{hu2017toward}, DeleteAndRetrieve \cite{li2018delete}, CycleRL \cite{xu2018unpaired}, StyleTransformer \cite{dai2019style}, Disentangled \cite{john2018disentangled}
MaskTransformer \cite{wu2020mask}, RACoLN \cite{lee2021enhancing}.

\subsection{Automatic Evaluation}
Target transferred sentences should be content-complete, fluent, and target-styled. Following previous work \cite{shen2017style, dai2019style, john2018disentangled, lee2021enhancing}, we compared three dimensions of generated samples: 1) Style control, 2) Content preservation, and 3) Fluency.

\noindent\textbf{Style Control} We measure style control automatically by evaluating the target sentiment accuracy of transferred sentences. We fine-tuned a pre-trained BERT \cite{devlin2018bert} with each dataset For an accurate evaluation of style control.

\noindent\textbf{Content preservation}
 To measure the content preservation, we adopt the BLEU (Bilingual Evaluation Understudy) \cite{papineni2002bleu} score as the evaluation metric. Specifically, we use the calculation tool provided by NLTK (Natural Language Toolkit) to calculate the BLEU score for the transfer sentence and the original sentence. A higher BLEU score indicates the transferred sentence can achieve better content preservation by retaining more words from the source sentence. If a human reference is available, we will calculate the BLEU score between the transferred sentence and corresponding reference as well. Two BLEU score metrics are referred to as self-BLEU and ref-BLEU respectively.

\noindent\textbf{Fluency}
A common method of evaluating fluency is calculating the complexity of the transfer sentence. Specifically, we use KenLM \cite{heafield2011kenlm} to train a 5-gram language model on the Yelp and IMDb datasets, and we use this model to calculate the perplexity of the sentence. Perplexity determines generation probability and fluency of a sentence. The lower the perplexity, the higher the generation probability.

\subsection{Implementation Details}
In this paper, we use a single layer Transformer with 4 attention heads in each layer as a Transformer block. The hidden size, embedding size, and positional encoding size in Transformer are all 256 dimensions. The size of bias and gain parameters of conditional layer norm is 256, either. The length of Normalizing Flow chain is 8. For balancing parameters of total loss function, we set to 0.5 for $\lambda_1$ and $\lambda_2$, and 1 for the rest.

\begin{table*}[ht]
\renewcommand\arraystretch{1.1}
\tabcolsep=3.2mm
\begin{tabular}{cccccccc}
\hline
\multicolumn{1}{l}{} & \multicolumn{4}{c}{Yelp}                                    & \multicolumn{3}{c}{IMDb}                      \\ \cline{2-8}
                     & ACC           & Self-BLEU     & ref-BLEU      & PPL         & ACC           & Self-BLEU     & PPL           \\ \cline{1-8}
Cross-Alignment      & 74.2          & 13.2          & 4.2           & 53          & 63.9          & 1.1           & \textbf{29.9} \\ 
ControlledGen        & 88.8          & 45.7          & 14.3          & 219         & 94.6          & 62.1          & 143           \\ 
DeleteAndRetrieve    & 87.7          & 29.1          & 10.4          & 60          & 58.8          & 55.4          & 57            \\ 
CycleRL              & 88.0          & 7.2           & 2.8           & 107         & \textbf{97.8} & 4.9           & 177           \\
MaskTransformer      & 91.8          & 54.6          & 19.3          & 81          & 95.7          & 63.9          & 92            \\
StyleTransformer     & 89.5          & 54.5          & 18.7          & 76          & 82.3          & 67.5          & 108           \\
RACoLN              & 91.3          & 59.4          & 20.0          & 60          & 83.1          & 70.9          & 45            \\
Disentangled         & 91.7          & 16.7          & 6.71          & \textbf{26} & N/A           & N/A           & N/A           \\ \cline{1-8}
StyleFlow            & \textbf{92.1} & \textbf{61.3} & \textbf{20.4} & 55          & 83.9          & \textbf{72.1} & 47   \\ \hline     
\end{tabular}
\caption{\label{Table2}
Automatic evaluation results on Yelp and IMDb datset.
}
\end{table*}

\subsection{Overall performance}
We compare our model with the baseline models, and the automatic evaluation result is presented in Table 1. Our
outperforms the baseline models in terms of content preservation on both of the datasets. Especially, on Yelp dataset, our model achieves 61.3 self-BLEU score, surpassing the previous state-of-the-art model by more than 2 points. Furthermore, our model also achieves the state-of-the-art result in content preservation on IMDB dataset (72.1 self-BLEU score), which is comprised of longer sequences than those of Yelp.

In terms of style transfer accuracy and fluency, our model is highly competitive. Our model achieves the highest score 92.1 on Yelp in style transfer accuracy. Additionally, our model shows the ability to produce fluent sentences as shown in the perplexity score.

With the automatic evaluation result, we see a trend of trade-off. Most of the baseline models are good at particular metric, but show space for improvement on other metrics. For example, Cross-Alignment constantly perform well in terms of perplexity, but their ability to transfer style and preserving content needs improvement. Disentangled is good at transferring style, but less so at preserving content. RACoLN achieves comparable performance across all evaluation metrics, but our model outperforms the model on most metric on both of the datasets. Consequently, the results show that our model is well-balanced, but also strong in every aspect of text style transfer.

\begin{table} \small 
\renewcommand\arraystretch{1.2}
\tabcolsep=0.4mm
\begin{tabular}{lcccc}
\hline
\centering
            
            & \multicolumn{1}{l}{\textbf{ACC}} & \multicolumn{1}{l}{\textbf{self-BLEU}} & \multicolumn{1}{l}{\textbf{ref-BLEU}} & \multicolumn{1}{l}{\textbf{PPL}} \\ \cline{1-5}
\multicolumn{1}{c}{StyleFlow}       & 92.1                             & 61.3                                   & 20.4                                  & 55                               \\ \hdashline
(-) attention coupling layer   & 89.6                             & 53.1                                   & 17.4                                  & 76                               \\
(-) data augmentation               & 90.5                             & 56.8                                   & 18.6                                  & 67                               \\
(-) \qquad CLN & 87.9                             & 57.2                                  & 18.3                                  & 62                               \\ \hline
  
\end{tabular}
\caption{\label{Table3}
Model ablation study results on Yelp dataset.
}
\end{table}

\subsection{Ablation Study}
In order to validate the proposed modules, we conduct ablation study on Yelp dataset which is presented in Table \ref{Table3}. To study the impact of different components on overall performance, we further did an ablation study for our model on Yelp dataset, and results are reported in Table \ref{Table3}. For better understanding the role of different methods, we remove or replace them to observe changes in model results. We observe a significant drop across all aspects without the attention-aware coupling layer. Especially the self-BLEU is reduced by more than 8 scores, which shows that the attention-aware coupling layer plays an essential role in content preservation. When we remove the data augmentation module, the model's effect and robustness are also reduced. In other case, where we remove the CLN and use style embedding as in the previous work \cite{dai2019style, yi2021text}, the model lowers the ability to transfer style feature, drop of around 4 score on ACC.

\subsection{Case Study}
To better understand the effect of different model in style transfer, we sampled several output sentences from the Yelp datset, which are shown in Table \ref{table 4}. When models transfer the style "positive to negative", StyleFlow generates more advanced vocabulary (e.g. "suffer in this horrible place", "know little about") to accurately express the "negative" meaning of sentences. Similarly, when models transfer the style "negative to positive", sentences generated by StyleFlow are more expressive and accurate, which indicates StyleFlow performs better than other models.

\tabcolsep=10mm
\begin{table*}[htbp]
\centering
\begin{tabular}{cl}
\hline
\multicolumn{2}{c}{\textbf{positive to negative}}                                                        \\ \hline
\textbf{Input}            & i will be going back and enjoying this great place !                         \\
\textbf{Cross-Alignment}  & i will be going to and enjoying this bad place !                             \\
\textbf{StyleTransformer} & i will be going back and dislike this bad place !                            \\
\textbf{RACoLN}           & i will not go back and hate this bad place !                                 \\
\textbf{StyleFlow(Ours)}  & I will not go back and \textcolor{red}{suffer in this horrible} place!                        \\
\textbf{Human}            & i won't be going back and suffering at this terrible place !                \\ \hline
\textbf{Input}            & awesome and fast service , these guys really know their stuff .              \\
\textbf{Cross-Alignment}  & worse and slow service , these guys really know their stuff .                \\
\textbf{StyleTransformer} & terrible and slow service , these guys really know their stuff .             \\
\textbf{RACoLN}           & \textcolor{red}{awful and slow service} , these guys really don't understand their stuff . \\
\textbf{StyleFlow(Ours)}  & horrible and slow service , these guys \textcolor{red}{know little about} their stuff .          \\
\textbf{Human}            & awful and slow service , these guys really don't know their stuff .         \\ \hline
\multicolumn{2}{c}{\textbf{negative to positive}}                                                         \\ \hline
\textbf{Input}            & the burgers were over cooked to the point the meat was crunchy .             \\
\textbf{Cross-Alignment}  & the burgers were less cooked to the point the meat was crunchy .             \\
\textbf{StyleTransformer} & the burgers were cooked to the point the meat was not crunchy .              \\
\textbf{RACoLN}           & the burgers were cooked well and the meat was delicious.                     \\
\textbf{StyleFlow(Ours)}  & the burgers were \textcolor{red}{ cooked to perfection} the meat was \textcolor{red}{delicious} .               \\
\textbf{Human}            & the burgers were cooked perfectly and the meat was juicy .                \\ \hline
\textbf{Input}            & it 's always busy and the restaurant is very dirty .                         \\
\textbf{Cross-Alignment}  & it 's not busy and the restaurant is very nice .                             \\
\textbf{StyleTransformer} & It's not always busy and the restaurant is clean.                            \\
\textbf{RACoLN}           & it 's always free and the restaurant is very  nice and clean .               \\
\textbf{StyleFlow(Ours)}  & it 's not always busy and the restaurant is  \textcolor{red}{very organized and clean} .        \\
\textbf{Human}            & it 's not always busy , and the restaurant is very clean .   \\ \hline              

\end{tabular}
\caption{\label{table 4}
Case study from Yelp dataset. The red words indicate good transfer.
}
\end{table*}

\section{Conclusion}
In this paper, we propose a novel disentanglement method StyleFlow to enhance content preservation in text style transfer tasks. StyleFlow adopts the Normalizing Flow structure to construct an attention-aware reversible encoder, which can encode in the forward process and decode in the reverse process losslessly. Due to its reversibility, StyleFlow can rebuild all transferred sentences back to their original style without losing any content in the process of cycle reconstruction. Furthermore, we can improve the robustness of our model with data augmentation through normalizing the flow. Experiment results demonstrate that our model preserves content well and achieves the state-of-the-art performance on the most metrics. In the future, we will study problems involving more than two styles and how to transfer multiple attributes of style to a target style.

\bibliography{anthology,custom}
\bibliographystyle{acl_natbib}

\end{document}